\documentclass[conference]{IEEEtran}
\IEEEoverridecommandlockouts
\usepackage{cite}
\usepackage{amsmath,amssymb,amsfonts}
\usepackage{url}
\usepackage{algorithm}
\usepackage[noend]{algpseudocode}
\usepackage{graphicx}
\usepackage{textcomp} 
\usepackage{xcolor}
\usepackage{cleveref}
\usepackage[utf8]{inputenc} 
\def\BibTeX{{\rm B\kern-.05em{\sc i\kern-.025em b}\kern-.08em
    T\kern-.1667em\lower.7ex\hbox{E}\kern-.125emX}}
\begin{document}

\title{Empowering Quality Diversity in Dungeon Design with Interactive Constrained MAP-Elites\\
{}
\thanks{The Evolutionary Dungeon Designer is part of the project \textit{The Evolutionary World Designer}, which is supported by The Crafoord Foundation.}
}

\author{\IEEEauthorblockN{Alberto Alvarez\IEEEauthorrefmark{1}\IEEEauthorrefmark{3}, Steve Dahlskog\IEEEauthorrefmark{1}\IEEEauthorrefmark{4}, Jose Font\IEEEauthorrefmark{1}\IEEEauthorrefmark{5}, and Julian Togelius\IEEEauthorrefmark{2}}
\IEEEauthorblockA{\IEEEauthorrefmark{1}Department of Computer Science and Media Technology\\
Malmö University, Sweden\\
Email: \IEEEauthorrefmark{3}alberto.alvarez@mau.se,
\IEEEauthorrefmark{4}steve.dahlskog@mau.se,
\IEEEauthorrefmark{5}jose.font@mau.se}
\IEEEauthorblockA{\IEEEauthorrefmark{2}Department of Computer Science and Engineering, Tandon School of Engineering\\
New York University, USA\\
Email: julian.togelius@nyu.edu}
}

\IEEEpubid{\begin{minipage}{\textwidth}\ \\[12pt]
978-1-7281-1884-0/19/\$31.00 \copyright 2019 IEEE \end{minipage}}

\maketitle

\begin{abstract}
We propose the use of quality-diversity algorithms for mixed-initiative game content generation. This idea is implemented as a new feature of the Evolutionary Dungeon Designer, a system for mixed-initiative design of the type of levels you typically find in computer role playing games. The feature uses the MAP-Elites algorithm, an illumination algorithm which divides the population into a number of cells depending on their values along several behavioral dimensions. Users can flexibly and dynamically choose relevant dimensions of variation, and incorporate suggestions produced by the algorithm in their map designs. At the same time, any modifications performed by the human feed back into MAP-Elites, and are used to generate further suggestions.
\end{abstract}

\begin{IEEEkeywords}
Procedural Content Generation, Evolutionary Algorithms, Mixed-Initiative Co-Creativity, Computer Games
\end{IEEEkeywords}

\section{Introduction}

Procedural Content Generation (PCG) refers to the generation of game content with none or limited human input~\cite{Yannakakis2018}, where game content could be anything from game rules, quests, and stories, to levels, maps, items, and music. While PCG has been present in some games since trailblazing games like \emph{Rogue}~\cite{michael_toy_1980} and \emph{Elite}~\cite{braben_elite_1984}, it has only been a popular academic research topic for a decade or so. Search-based PCG means using a global search algorithm such as an evolutionary algorithm to search content space~\cite{Togelius2011}.

Part of PCG's allure is the promise to produce game art and content faster and cheaper, as well as enabling innovative content creation processes such as player-adaptive games~\cite{shaker2012evolving, hastings_evolving_2009, dormansUnexplored2017}, data-driven content generation~\cite{Khalifa2018, Green2018}, and mixed-initiative co-creativity~\cite{Liapis2014}. Mixed-initiative co-creativity (MI-CC), a concept introduced by Yannakakis et al.~\cite{yannakakis2014micc}, refers to a creation process through which a computer and a human user feed and inspire each other in the form of iterative reciprocal stimuli. Some examples of this are \textit{Ropossum}~\cite{shaker2013ropossum}, \textit{Tanagra}~\cite{smith_tanagra:_2011}, \textit{CICERO}~\cite{Machado2017}, and \textit{Sentient Sketchbook}~\cite{liapis_generating_2013}. 

MI-CC aligns with the principles of lateral thinking and creative emotive reasoning: the processes of solving seemingly unsolvable problems or tackling non-trivial tasks through an indirect, non-linear, creative approach~\cite{Liapis2016}. Even more, MI-CC provides insight and understanding on the affordances and constraints of the human process for creating and designing games~\cite{Yannakakis2018}.

A key mechanism in MI-CC approaches is to present suggestions to players, and these suggestions must have high quality but also be sufficiently diverse. So-called quality-diversity algorithms~\cite{Pugh2016} are very well suited for this, as they find solutions that have high quality according to some measure, but are also diverse according other measures. MAP-Elites~\cite{Mouret2015} is a well-known algorithm of this type. Khalifa et al.~\cite{Khalifa2018} presented constrained MAP-Elites, a combination MAP-Elites with the feasible-infeasible concept from the FI2Pop genetic algorithm~\cite{Kimbrough2008}, and applied this to procedurally generating levels for bullet hell games.

The Evolutionary Dungeon Designer (EDD) is a MI-CC tool for generating dungeons for adventure games using a FI2Pop evolutionary approach~\cite{Alvarez2018, Alvarez2018a, Baldwin2017, Baldwin2017a}. This paper presents the Interactive Constrained MAP-Elites, an implementation of MAP-Elites into EDD's FI2Pop evolutionary algorithm, as well as introduces a continuous evolution process that takes advantage of MAP-Elites multidimensional discretization of the search space into cells. Results are analyzed and discussed regarding the improvements on quality diversity in the procedurally generated dungeons, as well as the effects of continuous evolution and dimension customization in a MI-CC approach.

\section{Background}
\subsection{Dungeons}
For more than 40 years, \emph{dungeons} have frequently been the setting for digital games and provided players with entertainment and excitement in particularly computer role-playing games (CRPGs) and adventure games. It seems that dungeons, as game settings, are as popular as ever, and shows no signs of going away~\cite{totten2017}. We can trace the first digital dungeons to the PLATO system back in 1975~\cite{barton08dad,brewer2016b} with games called ``\emph{pedit5}''~\cite{pedit5} and ``\emph{moria}''. Even though the layout of the dungeon in ``\emph{pedit5}'' was a fixed design, the game contained randomly generated encounters and rewards, making it a predecessor to the more commonly known \emph{Rogue}~\cite{michael_toy_1980} which provides the player with a new layout of the dungeon with every restart. However, prior to \emph{Rogue}, the game \emph{Beneath Apple Manor}~\cite{applemanor} made for the Apple contained a level generator which gave the player the possibility to replay the game with a different layout when starting the game. This feature of dungeons as ``randomized environments'' is the key element in so called Dungeon Hack games~\cite{batemanboon}. 

With regards to CRPGs and adventure games, it should be noted that they share the mechanisms of adventure and exploration whereas combat is more common in CRPG~\cite{rollings-adams}. Adventure games, on the other hand, more often contain puzzle solving as a mechanism. It is perhaps not that strange that dungeons are a popular setting for games in these genres, since they provide the following design elements: levels (several levels are needed with diverse layout and difficulty), collectibles (loot), boss fights, locked door and key (you need to find the key to open the door), wildcard enemies (placement, type and strength), monster generators (new monsters are generated until this mechanism is destroyed), and finally, exits and warps (which acts as transitions to other parts indicating progress in the game)~\cite{rollings-adams}. 

\begin{figure}[t]
\centerline{\includegraphics[width=9cm]{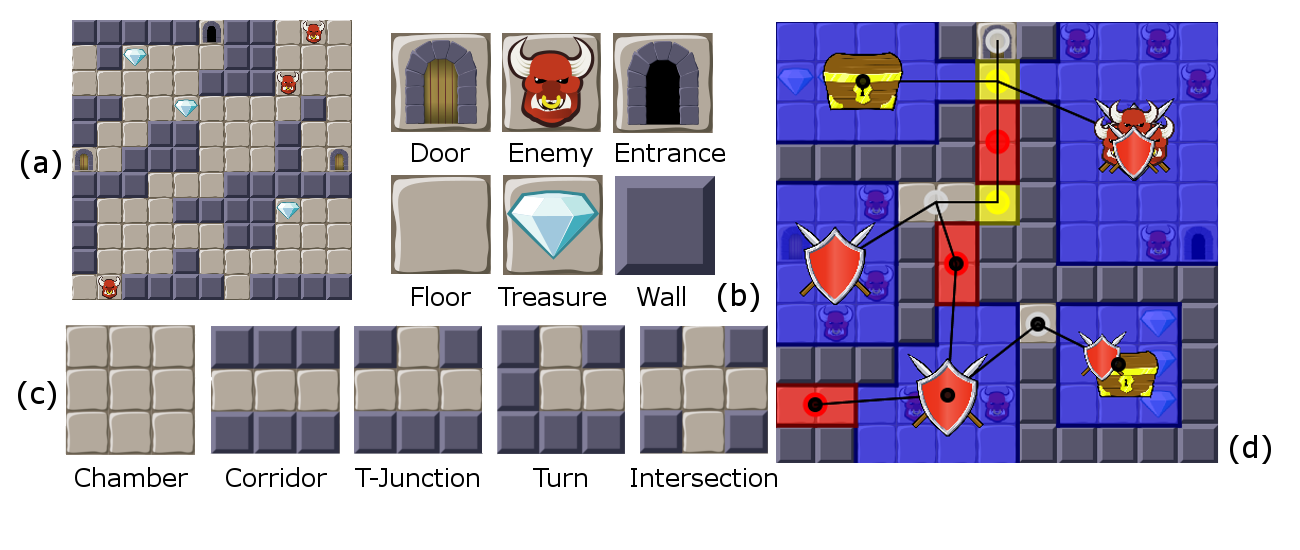}}
\caption{Main components in EDD. (a) Basic room, (b) different placeable tiles, (c) micro-patterns and (d) meso-patterns~\cite{Alvarez2018a}.}
\label{figs:basecomponents}
\end{figure}

\begin{figure}[t]
\centerline{\includegraphics[width=9cm]{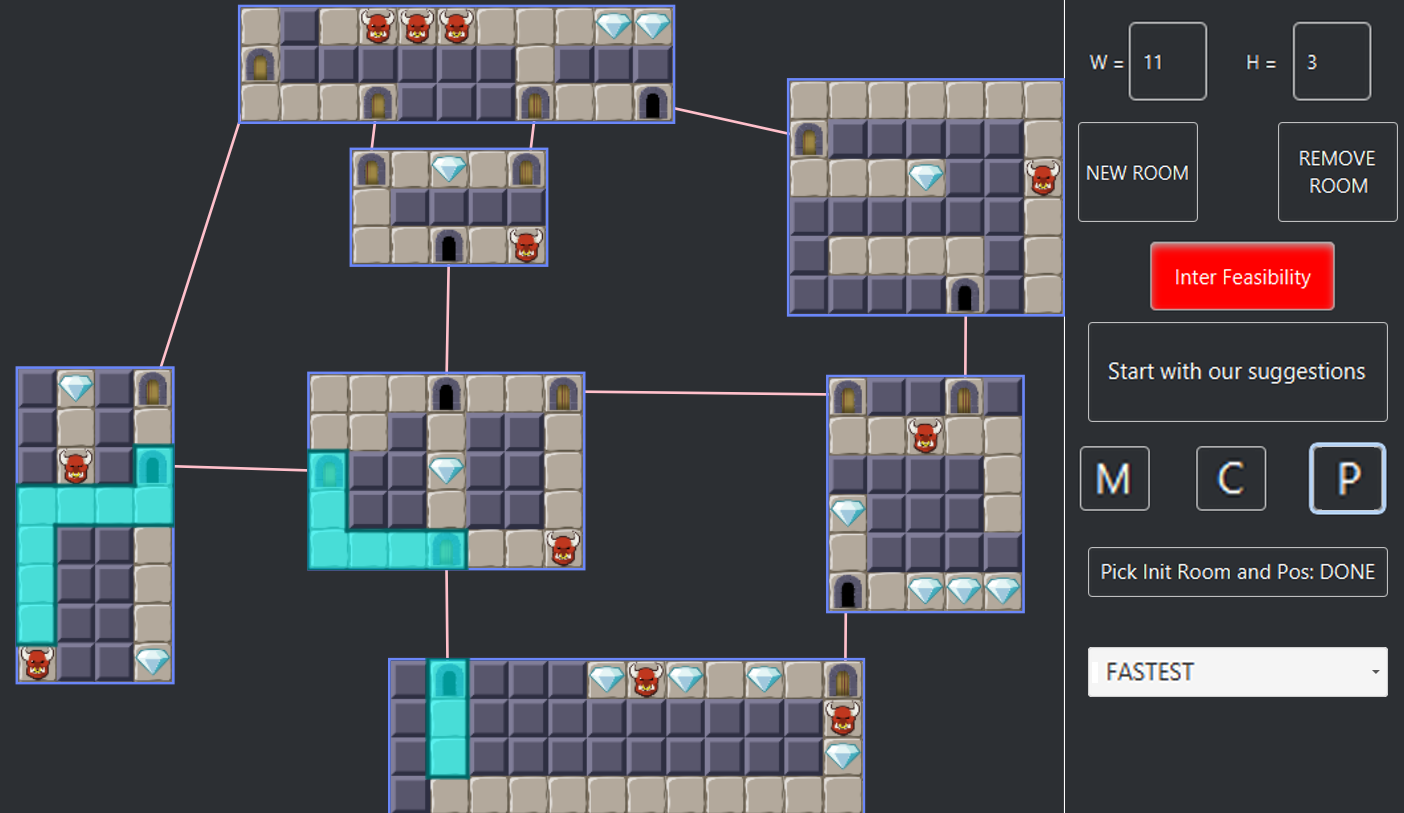}}
\caption{Screenshot of the dungeon editor screen in EDD, displaying a sample dungeon composed by seven rooms. The shortest path between two given tiles is highlighted in blue. The right pane contains all options for editing the dungeon. "M", "C", and "P" stand for "Move rooms", "Connect rooms", and "calculate Path".}
\label{figs:dungeonscreen}
\end{figure}

\subsection{Map-Elites for illuminating search spaces}

Quality-diversity algorithms are algorithms which search a space of solutions not just for the single best solution, but for a set of diverse solutions which are good. MAP-Elites maintains of map of good solutions~\cite{Mouret2015} and is perhaps the most well-known quality-diversity algorithm. The map is divided into a number of cells according to one or more feature dimensions (commonly, two dimensions are used). In each cell, a single solution is kept. At every update, an offspring is generated based on one or more existing solutions. That offspring is then assigned to a cell based on its feature dimensions, which might or might not be the same as the cell(s) its parent(s) occupy. 
If the new offspring has a higher fitness than the existing solution in that cell, it replaces the previous item in the cell. This process results in a map of solutions where each cell contains the best found solution for those particular feature dimensions.

\section{Evolving Dungeons as a Whole, Room by Room}

The Evolutionary Dungeon Designer (EDD) is a MI-CC tool that allows a human designer to create a 2D dungeon and its composing rooms (\Cref{figs:basecomponents}.a), being the designer able to manually edit both the dungeon - by placing and removing rooms - and the rooms - by separately editing the tiles (\Cref{figs:basecomponents}.b) that compose each room. EDD's underlying evolutionary algorithm provides procedurally generated suggestions, and is driven through the use of game design micro- and meso- patterns (\Cref{figs:basecomponents}.c and \Cref{figs:basecomponents}.d). A detailed description of all EDD's features can be found in~\cite{Baldwin2017a, Baldwin2017, Alvarez2018, Alvarez2018a}.

\begin{figure*}[t]
\centerline{\includegraphics[width=16cm]{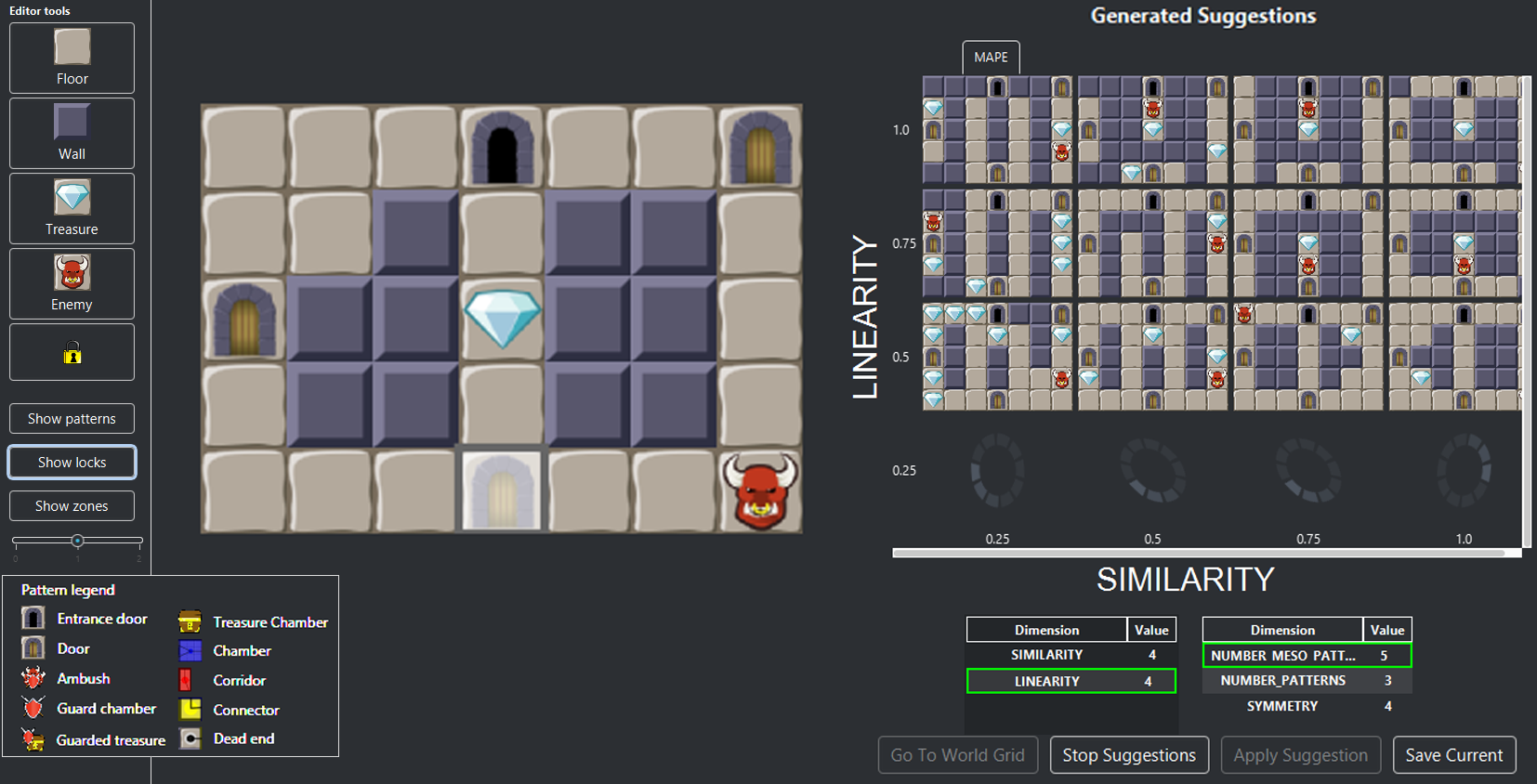}}
\caption{The room editor screen in EDD. The left pane contains all the options for manually editing the room displayed at the center-left of the screen. The right section displays the procedurally generated suggestions.}
\label{figs:roomscreen}
\end{figure*}

This section presents the latest version of EDD\footnote{available for download at \url{https://drive.google.com/file/d/1lCUfc4OF7lY3vUlPzAqf7i7OUfaKQoem/view}}, which includes significant improvements based on the outcomes from the qualitative analysis discussed in~\cite{Alvarez2018}. The most significant upgrade is replacing the grid-based backbone that represented the dungeon by a more flexible graph-based representation. A dungeon is now a graph of interconnected rooms of any given size between $3\times3$ and $20\times20$ tiles. The smallest allowed dungeon is composed by two rooms connected once to each other. The designer can perform the following new actions: 

\begin{itemize}
\item adding disconnected rooms to the dungeon. Rooms may also be removed at any time.
\item connecting any pair of rooms by adding a new bi-directional connection to the graph. Rooms interconnect from and to passable border tiles (self-loops are not allowed). Both ends are marked with a door tile (\Cref{figs:basecomponents}.b). A single border tile can only hold one connection, implying that a room can have as many connections as passable border tiles. Connections and rooms can be removed at any time, and their associated doors removed with them.
\item calculating paths between any pair of passable tiles located in any connected room. Paths are automatically calculated following one the following heuristics: \textit{fastest} returns the shortest path, \textit{rewarding} returns the path that traverses the highest number of treasure tiles, \textit{less danger} provides a path with the fewest number of enemies, whereas \textit{more danger} does the opposite. 
\end{itemize}

The designer is required to select one of the added rooms as the \textit{initial room}, which is the room used by the player to enter the dungeon (for the first time). This selection can be modified unlimited times. The \textit{initial room} is used by EDD to calculate the feasibility of the dungeon. A dungeon is considered feasible when there is at least one path between the \textit{initial room} and any other passable tile in every room. Rooms and doors that aren't reachable from the \textit{initial room} are highlighted in red, so that they can be easily identified by the designer. This feasibility constraint ensures that all passable tiles are accessible, avoiding the possibility of accidentally creating unreachable areas.  

\subsection{The mixed-initiative workflow in EDD}

The starting screen in EDD is the dungeon editor screen, shown in \Cref{figs:dungeonscreen}. Every new room is empty (composed solely of floor tiles) when created and is placed detached from the dungeon graph. After manually connecting the room to the dungeon with at least one connection, the designer has the option to populate the room using the room editor screen (\Cref{figs:roomscreen}). This screen can be reached in two different ways:

\begin{enumerate}
\item directly: by double-clicking or zooming in (by using the mouse steering wheel or by pinching on the touchpad) on the room. 
\item indirectly: by clicking on the "Start with our suggestions" button on the right pane (\Cref{figs:dungeonscreen}), six procedurally generated suggestions are displayed on a separate screen. The selected suggestion is then opened in the room editor screen. 
\end{enumerate}

\Cref{figs:roomscreen} shows the room editor screen displaying a sample room with the dimensions 7x5 tiles. The left pane lists all the available options for manually editing the room. Manual editing is carried out by brush painting over the room with one of the available tile types: floor, wall, treasure, or enemy. There are two brush sizes (single tile, and five-tile cross shape), and control-clicking allows the designer to bucket paint all adjacent tiles of the same type. Brush painting with the lock button on preserves selected tiles in all the procedurally generated suggestions. A detailed description of all the options in this pane is included in~\cite{Alvarez2018, Alvarez2018a}.

The right side of the screen displays the procedurally generated suggestions, by means of the Interactive Constrained MAP-Elites genetic algorithm (\Cref{section:illuminating}). The "Generate/Stop Suggestions" button at the bottom toggles this algorithm on and off. Once started, the algorithm continuously populates the suggestions pane with new optimal individuals. The evolutionary process is fed with the manually edited room, so that every change in the room affects the generated suggestions. By clicking on "Apply Suggestion", the manually edited room is replaced by the selected suggestion, thus affecting the upcoming procedural suggestions. "Go To World Grid" takes the user back to the dungeon editor screen.

\section{Interactive Constrained MAP-Elites\label{section:illuminating}}


EDD uses a single-objective fitness function with a FI2Pop genetic algorithm where fitness is a weighted sum divided equally between (1) the inventorial aspect of the rooms, which relates to the placement of enemies and treasures in relation to doors and target ratios, and (2) the spatial distribution of the design patterns, which relates to the distribution between corridors and rooms, and the meso-patterns that those encompass. An in-depth explanation of EDD's fitness function can be found in~\cite{Alvarez2018a, Baldwin2017}.

The overarching goal of MI-CC is to collaborate with the user to produce content, either to optimize (i.e. exploit) their current design towards some goal or to foster (i.e. explore) their creativity by surprising them with diverse proposals. By implementing MAP-Elites~\cite{Mouret2015} and continuous evolution into EDD, our algorithm can (1) account for the many dimensions that a user can be interested, (2) explore multiple areas of the search space and produce a diverse amount of high-quality suggestions to the user, and (3) still evaluate how interesting and useful the tile distribution is within a specific room. Henceforth, we name the presented approach \textbf{Interactive Constrained MAP-Elites} (IC MAP-Elites). 

\subsection{Illuminating Dungeon Populations with MAP-Elites}

MAP-Elites explores the search space more vastly by separating certain interesting dimensions, that affect different aspects of the room such as playability or visual aesthetics, from the fitness function, using them to categorize rooms into niches (cells). 

\subsubsection{Dimensions}

Dimensions in MAP-Elites are identified as those aspects of the individuals that can be calculated in the behavioral space, and that are independent of the fitness calculation. 
EDD offers the designer the possibility to choose among the following dimensions, two at a time:

\textbf{Symmetry and Similarity.} We choose Symmetry as a consideration of the aesthetic aspects of the edited room since symmetric structures tend to be more visually pleasing. Similarity is used to present the user variations of their design but still preserving their aesthetical edits. Symmetry is evaluated along the X and Y axes, backslash and front slash diagonal and the highest value is used as to how symmetric a room is. Similarity is calculated through comparing tile by tile with the target room. Formulas, information and support for both evaluations are explained in greater details at~\cite{Alvarez2018a}, where both of them were used as aesthetic fitness evaluations.

\textbf{Number of Meso-patterns.} The number of meso-patterns correlates to the type and amount of encounters the designer wants the user to have in the room in a more ordered manner. The considered patterns are the treasure room (tr), guard rooms (gr), and ambushes (amb). Meso-patterns associate utility to a set of tiles in the room, for instance, a long chamber filled with enemies and treasures could be divided into 2 chambers, the first one with enemies and the second one with treasures so the risk-reward encounter is more understandable for the player. Since we already analyze the rooms for all possible patterns, the number of meso-patterns is simply $\#MesoPat=tr, gr, amb \in AllPatterns$. Equation (\ref{eq:meso-pat-eq}) presents the dimensional value, and since the used meso-patterns can only exist in a chamber, we normalize by the maximum amount of chambers in a room, which are of a minimum size of $3x3$, and results in $Max_{chambers}=\left\lfloor Cols/3 \right\rfloor * \left\lfloor Rows/3 \right\rfloor$.



\begin{equation} \label{eq:meso-pat-eq}
D_{mesoPat} = \min \left\{ \dfrac{\#MesoPat}{Max_{chambers}}, 1.0 \right\}
\end{equation}

\textbf{Number of Spatial-patterns.} By spatial-patterns we mean chambers (c), corridors (cor), connectors (con), and nothing (n). We identify the number of spatial-pattern relates to how individual tiles group (or not) together to form spatial structures in the room. The higher the amount of spatial-patterns the lesser tiles will be group together in favor of more individualism. For instance, a room with one spatial-pattern can be one with no walls and just an open chamber, while a room with a higher number of spatial-patterns would subdivide the space with walls, using tiles for more specific patterns. Equation (\ref{eq:spatial-pat-eq}) presents how we calculate the value for such a dimension. The number of spatial patterns is simply $\#SpatialPat=c, n, cor, con \in AllPatterns$, we then normalize it by the largest side of the room and multiply it by a constant value, determined as $K=4.0$ through a process of experimentation since it resulted in a good estimation of the amount of spatial patterns in the room.




\begin{equation} 
\label{eq:spatial-pat-eq}
D_{spatialPat} = \min\left\{\frac{\#SpatialPat}{\max\left\{{Cols, Rows}\right\} * \textit{K}}, 1.0\right\}
\end{equation}

\textbf{Linearity.} Linearity represents the number of paths that exist between the doors in the room. This relates to the type of gameplay the designer would like the room to have by the distributions of walls among the room. Having high linearity in a room does not need to only be by having a narrow corridor between doors but could also be generated by having all doors in the same open space (i.e. the user would not need to traverse other areas) or by simply disconnecting all paths between doors. Equation (\ref{eq:Linearity-eq}) shows the linearity calculation. Due to the use of patterns, we calculate the paths between doors as the number of paths that exist from a spatial-pattern containing a door to another. Finally, this is normalized by the number of spatial patterns in combination with the number of doors and their possible neighbors.

\begin{equation} \label{eq:Linearity-eq}
D_{lin} = 1 \text{--} \frac{AllPathsBetweenDoors} {\#spatialPat + \#NeighborsPerDoor}
\end{equation}

\subsubsection{Continuous Evolution}



EDD implements continuous evolution in two ways. First, the EA constantly updates the target room and configuration with the most recent version of the user’s design, and once the suggestions are broadcasted, that room is incorporated without changes to the population of individuals in the corresponding cell. Secondly, by changing the dimension information and their granularity for the MAP-Elites, which can be done at any given time by the designer. 

Provided that EDD already uses a FI2Pop, we took as a starting point the constrained MAP-Elites presented by Khalifa et al.~\cite{Khalifa2018}, where the illuminating capabilities of MAP-Elites explore the search space with the constraints aspects of FI2Pop. This approach manages two different populations within each cell, a feasible and an infeasible one. Individuals move across cells when their dimension values change, or between the feasible and infeasible population according to their fulfillment of the feasibility constraint.

\algnewcommand\algorithmicforeach{\textbf{for each}}
\algdef{S}[FOR]{ForEach}[1]{\algorithmicforeach\ #1\ \algorithmicdo}

\algblockdefx{MRepeat}{EndRepeat}{\textbf{repeat}}{}
\algnotext{EndRepeat}

\begin{algorithm}
\caption{Interactive Constrained MAP-Elites}\label{alg:IC-MAPE}
\begin{algorithmic}[1]
\Procedure{IC-MAP-Elites($\protect[\{d_1,v_1\},...,\{d_n,v_n\}]$)}{}
\State $target \gets curEditRoom$ \Comment{Always in background}
\State createCells$(\protect[\{d_1,v_1\},...,\{d_n,v_n\}])$
\For{$i \gets 1$ to $PopSize$} 
     \State add mutate$(target)$ to $population$
\EndFor
\State CheckAndAssignToCell$(population)$ 
\While {true} \Comment{start continouous evo}
    \For{$generation \gets 1$ to $publishGen$}
        \If {$\textit{dimensionsChanged}$}
            \State $previousPop \gets cells_{pop}$
            \State createCells$(newDimensions)$
            \State checkAndAssignToCell$(previousPop)$ 
        \EndIf
        \MRepeat{ \text{[for feasible \& infeasible pop.]}}
            \For{$i \gets 1$ to $ParentIteration$}
                \State $curCell \gets \text{rndCell}(cells)$
                \State add tournament$(curCell)$ to $parent$
            \EndFor
            \State $offspring \gets  \text{crossover}(Parent)$
            \State checkAndAssignToCell$(offspring)$
        \EndRepeat
        \State sortAndTrim$(cells)$
    \EndFor
    \State broadcastElites() \Comment{render elites}
    \State $pop' \gets cells_{population}$
    \State add mutate$(cells_{pop})$ to $pop'$
    \State add $target$ to $pop'$
    \State checkAndAssignToCell $(pop')$
    \State sortAndTrim$(cells)$
\EndWhile
\EndProcedure
\Procedure{createCells(dimensions)}{}
    \ForEach{$dim \in dimensions $}
        \State add newCell$(dim_d, dim_v)$ to $cells$
    \EndFor
\EndProcedure
\Procedure{$\protect \text{check\&AssignToCell}(curPopulation)$}{}
    \ForEach{$individual \in curPopulation $}
        \State $individual_f \gets evaluate(individual)$ 
        \State $individual_d \gets dim(individual)$
        \State add $individual$ to $cell_{pop}(individual_d)$
    \EndFor
\EndProcedure
\end{algorithmic}
\end{algorithm}

\begin{figure*}[t]
\centerline{\includegraphics[width=18cm]{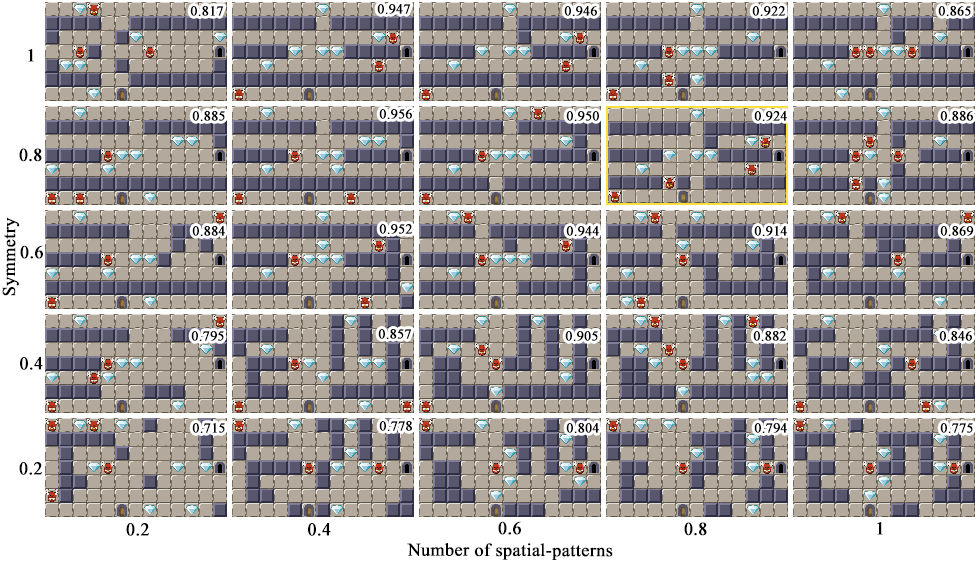}}
\caption{Rooms at generation $2090$ targeting Number of spatial-patterns (X) and Symmetry (Y). Each cell displays (top-right) the fitness of the optimal individual in its related feasible population. }
\label{figs:patt_sym}
\end{figure*}

\subsubsection{Algorithm}

The current evolutionary algorithm is depicted in Algorithm \ref{alg:IC-MAPE}. Cells are first created based on the dimensions selected by the user and proceed to initialize the population based on the user's design, evaluate it and assign each individual to the corresponding cell. Before starting each generation, we check if the dimensions have changed, and if so, recreate the cells and populate them with the previous individuals, and proceed through the evolutionary strategies. Selection is through tournament with a random number of competing parents and offspring are produced through a two-point uniform crossover with a chance of mutation. Offspring are placed in the correct cell and population after calculating their fitness and dimension's information. Finally, cells eliminate the low-performing individuals that over-cap their maximum capacity. Since interbreeding is not allowed, and can only happen indirectly (i.e. the offspring changing population and then used for breeding in consequent generations), the strategies are repeated for each of the population.

This procedure is repeated until the user decides to stop the algorithm. Meanwhile, the EA runs for $n$ generations, and once it reaches the specified limit, it broadcasts the found elites. In order to push the exploration, we first mutate all the individuals from all the populations and cells (while retaining the previous population), and add them into the same pool together with the current edited room without changes. Finally, we evaluate and assign all the individuals to the correct cells, and cells that are over maximum capacity eliminates low-performing individuals.

\section{Experiments}

We ran a set of experiments to test the results from the IC MAP-Elites using all possible combinations of the five available dimensions using two dimensions at a time. All experiments were run using $13\times7$ rooms, the same room size as in \emph{The Binding of Isaac}~\cite{mcmillen_binding_2011}, a representative example of a dungeon based adventure game.
In each experiment, the initial population was set to $1000$ mutated individuals distributed in feasible and infeasible populations in all cells which were set to a maximum capacity of $25$ individuals each. The EA ran continuously, every $100$ generations rendered the most prominent cells, and at each of the generations, it selected $5$ parents per population among the different cells. Offsprings were produced through a two-point crossover, and were mutated with a 30\% chance.  

\Cref{section:results} describes the results achieved and analyzes them in terms of the quality diversity of the suggestions obtained, the existing correlations found between each pair of dimensions, as well as the effects of integrating the MAP-Elites approach into a continuously evolving environment.

\section{Results and Discussion\label{section:results}}
\Cref{figs:patt_sym} shows a grid containing the best found suggestions at generation $2090$, while aiming for number of spatial-patterns at the X-axis and symmetry at the Y-axis with a granularity of $5$. Each cell displays the optimal individual of the feasible population under a given pair of dimension values. The fitness score is displayed on the cells' top-right corner.

The fitness evaluation in IC MAP-Elites is quite lightweight in terms of computational cost, so that the grid of suggestions is completed in a matter of seconds. This is of key importance for successfully implementing continuous evolution, so that the influence of each manual change in the edited rooms is reflected in the suggestions almost instantly. The feeling of immediacy is further increased through updating cells as soon as a new optimal individual is produced and incorporated to the cell’s underlying feasible population.

\begin{figure}[ht!]
\centerline{\includegraphics[width=8cm]{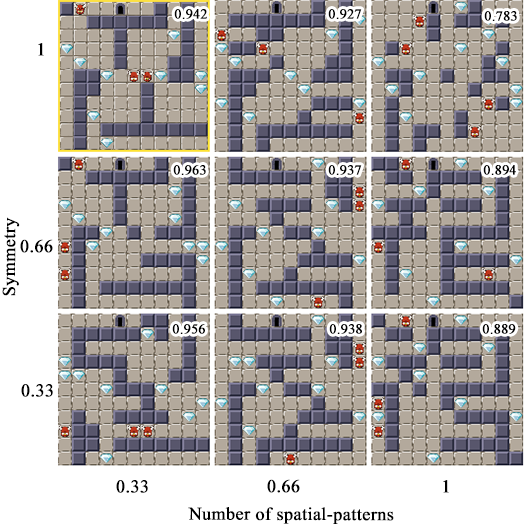}}
\caption{Rooms at generation $5303$ targeting the same dimensions as in \Cref{figs:patt_sym}, but with the size $11\times11$ instead.}
\label{figs:patt_sym3}
\end{figure}

\begin{figure*}[ht]
\centerline{\includegraphics[width=18cm]{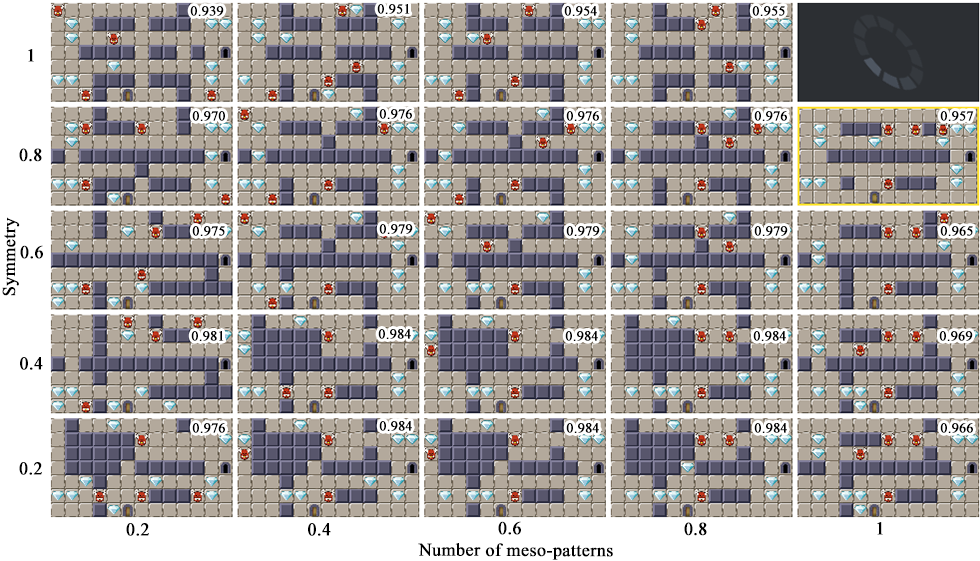}}
\caption{Rooms at generation $7088$ targeting Number of meso-patterns at the X-axis and Symmetry at the Y-axis. The top-right cell shows that no optimal room could be generated under dimension values $[1,1]$. }
\label{figs:meso_sym}
\end{figure*}

Results in \Cref{figs:patt_sym} are representative of the good quality diversity solutions produced by EDD. The average fitness across cells is $0.872$, and the highest fitness is $0.956$ (cell $[0.4,0.8]$). No two rooms are the same. As intended, high levels of symmetry are displayed in the upper rows, gradually decreasing towards the bottom row. Similarly, rooms in the leftmost column contain lower amounts of spatial patterns, increasing towards the rightmost column. Lower amounts of spatial patterns translate into more open rooms with almost no corridors and one or two large adjacent chambers (as in cell $[0.2, 0.2]$), as opposed to highly pattern filled rooms that comprise intricate pathways converging at one or two small chambers (cell $[1, 0.2]$). Fitness values show that some dimension combinations are harder to optimize than others, so that the whole grid depicts a gradient landscape of the compatibility between each pair of dimensions. 

The bottom-left corner in \Cref{figs:patt_sym} shows difficulties producing symmetric rooms with low amounts of spatial patterns, as opposed to rooms with many corridors (upper-right corner), which seem to favor the generation of symmetrical structures. The bottom row shows that aiming for low symmetry generally produces slightly less optimal results, whereas the top row shows that corridors are the most favorable spatial pattern for building symmetric rectangular rooms. Additional experiments (\Cref{figs:patt_sym3}) show that medium-large square rooms favor the appearance of chambers in combination with corridors for achieving symmetric rooms, thus revealing that squareness and size are important factors for the appearance of chambers in symmetric rooms.

\Cref{figs:meso_sym} contains the rooms generated at generation $7088$ while targeting number of meso-patterns at the X-axis and symmetry at the Y-axis. The top-right cell is empty because its related feasible and infeasible populations are empty, that is, no individuals with value $1$ for both dimensions have been found. The number of empty cells in the earlier generations $3722$, $3875$, and $5864$ were $8$, $7$, and $2$, respectively, indicating that some dimensional values for meso-patterns and symmetry take longer to converge. The continuous nature of IC MAP-Elites fills out the initially empty cells while the designer works with the already generated suggestions. The right half of the grid shows that a combination of small chambers and short corridors favors the appearance of multiple meso-patterns, such as treasure chambers, guarded chambers, and ambushes.

\begin{figure}[ht!]
\centerline{\includegraphics[width=8cm]{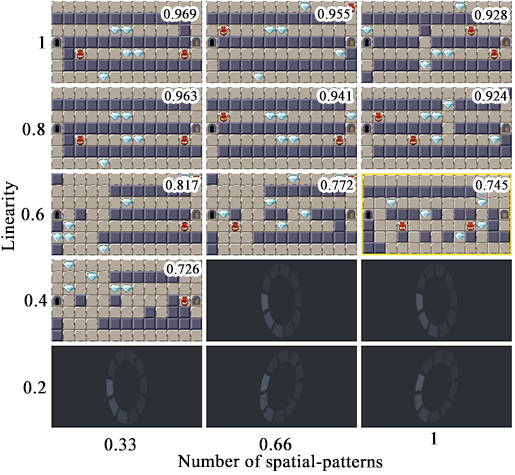}}
\caption{Rooms at generation $12545$ targeting Number of spatial-patterns at the X-axis and Linearity at the Y-axis.}
\label{figs:lin_patt}
\end{figure}

\begin{figure}[ht!]
\centerline{\includegraphics[width=8cm]{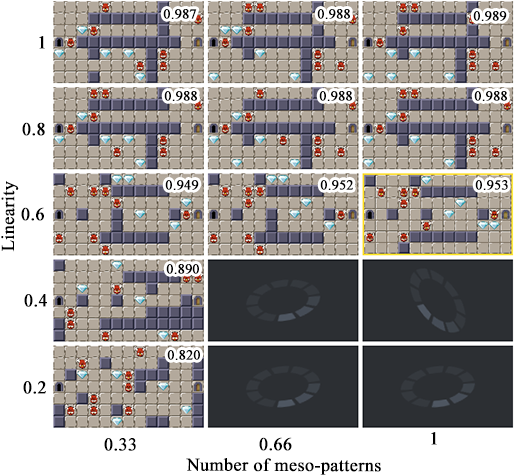}}
\caption{Rooms at generation $20348$ targeting Number of meso-patterns at the X-axis and Linearity at the Y-axis.}
\label{figs:lin_meso}
\end{figure}

\Cref{figs:lin_patt,figs:lin_meso} show how low valued linearity does not cope well with neither spatial- nor meso-patterns. High linearity tends to create one single pathway, either one long corridor or a wide chamber, that connects the doors in the room. Low linearity results in the opposite, scattering multiple small passages that increase the connectivity between doors but do not count neither as spatial- nor as meso-patterns.

Due to its nature, the performance of similarity in combination with other dimensions has been found to be very dependant on the characteristics already present in the manually edited room. I.e., if this room is already highly symmetric, EDD has problems at preserving similarity while targeting low values of symmetry. This behavior is reported when combining similarity with the other dimensions.

\section{Conclusions and Future Work\label{section:conclusion}}
We have presented the Interactive Constrained MAP-Elites, a continuous implementation of MAP-Elites into the Evolutionary Dungeon Designer, creating a MI-CC tool where the users influence the EA through their design, as well as by choosing which dimensions to explore and the granularity of such. 

The presented approach allows the designer to have a fast interaction with the EA through re-targeting and re-scaling the dimensions at will and at any moment. The continuous evolution fits perfectly to the mixed-initiative approach, providing a dynamic search that reacts on the fly to the different interactions of the user, as well as constantly offering new suggestions accordingly. Moreover, mixed-initiative fills the lapses between generations by inviting the designer to permeate the suggestions with custom aesthetics, challenges, paths, and other design decisions. Results show that this approach creates a very fluent workflow of mutual inspiration between designer and tool, yet offering highly customized quality diversity procedural suggestions. 

Results also allowed us to study the compatibility between each pair of dimensions, spotting existing correlations among them and with the fitness function, as well as compatibility pitfalls that leave room for further analysis.

We aim to validate IC MAP-Elites with a user study, as well as to explore alternatives to visualize higher dimensions through the use of CVT-MAP-Elites~\cite{cvt-mape2016} and Cluster MAP-Elites~\cite{cluster-mape2017}, analyze the effect of including more dimensions, and performing agent-based dungeon evaluation to improve the fitness calculation by incorporating automatic gameplay data.

\bibliographystyle{unsrt}

\vspace{12pt}

\end{document}